\documentclass[nohyperref]{article}
\usepackage{microtype}
\usepackage{graphicx}
\usepackage{subfigure}
\usepackage{caption}
\usepackage[font=small,skip=0pt]{caption}
\usepackage{booktabs}
\usepackage{hyperref}

\usepackage[accepted]{icml2022}
\usepackage{array}
\usepackage{amssymb}
\usepackage{mathtools}
\usepackage{amsfonts}
\usepackage{amsmath}
\usepackage{multicol}
\usepackage{tikz}
\usepackage{siunitx}
\usepackage{makecell}

\mathtoolsset{showonlyrefs}
\usepackage{multirow}
\usepackage[nodisplayskipstretch]{setspace}
\newcommand{\mll}{<\!\!\!<} 

\icmltitlerunning{LeNSE: Learning To Navigate Subgraph Embeddings for Large-Scale Combinatorial Optimisation}

\begin{document}

\twocolumn[
\icmltitle{LeNSE: Learning To Navigate Subgraph Embeddings \\
           for Large-Scale Combinatorial Optimisation}


\icmlsetsymbol{equal}{*}

\begin{icmlauthorlist}
\icmlauthor{David Ireland}{dept}
\icmlauthor{Giovanni Montana}{stats,dept}
\end{icmlauthorlist}

\icmlaffiliation{dept}{Warwick Manufacturing Group, University of Warwick, Coventry, United Kingdom}
\icmlaffiliation{stats}{Department of Statistics, University of Warwick, Coventry, United Kingdom}

\icmlcorrespondingauthor{Giovanni Montana}{g.montana@warwick.ac.uk}

\icmlkeywords{Reinforcement Learning, Deep Reinforcement Learning, Combinatorial Optimization, ICML}

\vskip 0.3in
]

\printAffiliationsAndNotice{}

\begin{abstract}
  
Combinatorial Optimisation problems arise in several application domains and are often formulated in terms of graphs. Many of these problems are NP-hard, but exact solutions are not always needed. Several heuristics have been developed to provide near-optimal solutions; however, they do not typically scale well with the size of the graph. We propose a low-complexity approach for identifying a (possibly much smaller) subgraph of the original graph where the heuristics can be run in reasonable time and with a high likelihood of finding a global near-optimal solution. The core component of our approach is LeNSE, a reinforcement learning algorithm that learns how to navigate the space of possible subgraphs using an Euclidean subgraph embedding as its map. To solve CO problems, LeNSE is provided with a discriminative embedding trained using any existing heuristics using only on a small portion of the original graph. When tested on three problems (vertex cover, max-cut and influence maximisation) using real graphs with up to $10$ million edges, LeNSE identifies small subgraphs yielding solutions comparable to those found by running the heuristics on the entire graph, but at a fraction of the total run time. Code for the experiments is available in the public GitHub repo at \url{https://github.com/davidireland3/LeNSE}.

\end{abstract}

\section{Introduction}
    
    Combinatorial optimisation (CO) problems involve the search for maxima or minima of an objective function whose domain is a discrete but large configuration space \cite{wolsey1999integer, korte2011combinatorial, grotschel2012geometric}.  Most CO problems can be formulated naturally in terms of graphs  \cite{avis2005graph, arumugam2016handbook, vesselinova2020learning}, and an optimal solution  typically consists of a set of  vertices that optimises the objective function.  Some well-known examples are the Vertex Cover Problem (VCP) \cite{dinur2005hardness}, i.e. the problem of finding a set of vertices that includes at least one endpoint of every edge of the graph, and the Max-Cut (MC) problem \cite{goemans1995improved}, i.e. finding a cut whose size is at least the size of any other cut. Many CO problems have a wide range of real-world applications, e.g. in biology \cite{wilder2018optimizing, reis2019simultaneous}, social networks \cite{brown1987social, valente1996social, rogers2010diffusion, chaoji2012recommendations},  circuit design \cite{barahona1988application} and auctions \cite{kempe2010frugal, dobzinski2011mechanisms, amanatidis2017budget}.

    Finding the optimal solution to a CO problem requires an exhaustive search of the solution space; with many canonical CO problems being NP-hard \cite{karp1972reducibility}, this usually means that they are unsolvable in practice. However,  exact optimal solutions are often not required, and many heuristic approaches have been developed to obtain near-optimal solutions in some reasonable time \cite{hochbaum1982approximation, goemans1995improved, applegate2001tsp, kempe2003maximizing, karakostas2005better, ausiello2012complexity}. More frequently, the practical application of these algorithms involves very large graphs. This significantly increases the run time of the heuristics and, in some case, may preclude the use of such algorithms altogether, e.g. due to memory constraints.  
    
    For instance, the Influence Maximization (IM) problem consists of finding a seed set composed of vertices that maximize their influence spread over a social network, which may have millions of vertices and/or edges. A greedy algorithm proposed by \citet{kempe2003maximizing} requires the objective function to be evaluated at every vertex in the graph; this is expensive as the expected spread of a vertex is \#P-hard to evaluate \cite{kempe2003maximizing, chen2010scalable}. Consequently, a large body of work has been carried out to develop more scalable algorithms \cite{goyal2011data, cheng2013staticgreedy, cheng2014imrank, cohen2014sketch, borgs2014maximizing, tang2014influence, tang2015influence}. The current state of the art algorithm for solving IM, the Influence Maximization Martingale (IMM) algorithm \cite{tang2015influence}, is also known to scale poorly to large graph instances due to its memory requirements \cite{arora2017debunking}. 
        
    In this work we present a general framework, based on supervised and reinforcement learning, for leveraging heuristics which have been crafted with extensive domain knowledge but may not scale well to large problem instances. Our proposed algorithm, LeNSE (\textbf{Le}arning to \textbf{N}avigate \textbf{S}ubgraph \textbf{E}mbeddings), learns how to prune a graph, significantly reducing the size of the problem by removing vertices and edges so that the heuristic can find a nearly-optimal solution at a fraction of the time that would have been taken when using the entire graph. The motivation for our approach comes from the fact that a subgraph is an easier target to identify, compared to the exact solution; instead of finding a needle in a haystack, we instead learn how to remove parts of the haystack where the needle is unlikely to be. 
     
    The graph pruning process is framed as a sequential decision making problem, which is solved using Q-learning \cite{watkins1992q}. Starting with any randomly chosen subgraph of fixed size, LeNSE sequentially modifies the current subgraph to a slightly altered one, and repeats this process until the current subgraph is deemed highly likely to contain a nearly-optimal solution. To efficiently navigate the space of possible subgraphs so that the highest quality one is reached in the fewest number of steps, LeNSE relies on a discriminative subgraph embedding as its \emph{navigation map}. To learn this embedding, we make use of recent advances in geometric deep learning, and specifically graph neural networks (GNNs) with graph coarsening layers \cite{ying2018hierarchical, cangea2018towards, lee2019self, liang2020mxpool}. By construction, the position of an embedded subgraph reflects its predicted quality. LeNSE makes use of these quality estimates to learn optimal graph modifications that progressively move the subgraph towards more promising regions. 
    
    Experimentally we test LeNSE on three well-know CO problems using established heuristics. Our extensive experimental results demonstrate that LeNSE is capable of removing large parts of the graph, in terms of both vertices and edges, without any significant degradation to the performance compared to  performing no pruning at all, but at a fraction of the overall run time. We also compare LeNSE to two baselines, a GNN vertex classifier and a recently introduced pruning approach \cite{manchanda2020gcomb}. Remarkably, we show that all training can be done on some small, random sample of the problem graph whilst being able to scale up to the larger test graph at inference time. This means that all expensive computations needed to train LeNSE, such as running the heuristic to obtain a solution, are only ever done on small graphs. We also show that using LeNSE for the IM problem on a large graph give speed ups of more than 140 times compared to not doing any pruning. 

\section{Related work}

   In recent years, many machine learning based solutions have been investigated to solve CO problems \cite{bengio2020machine, vesselinova2020learning, mazyavkina2021reinforcement}. Recent contributions in the field have started to leverage techniques from geometric deep learning, which is concerned with the application of neural network techniques to non-Euclidean domains \cite{bronstein2017geometric, battaglia2018relational, wu2020comprehensive, zhou2020graph}. These developments have started to play an important role in learning based CO algorithms as they capture and exploit the graphical nature of problems. Many methods learn vertex/edge embeddings \cite{kipf2016semi, hamilton2017inductive, gilmer2017neural, velivckovic2017graph, velivckovic2018deep, brody2021attentive}, whilst more recently methods have been explored to obtain graph level embeddings \cite{ying2018hierarchical, cangea2018towards, lee2019self, liang2020mxpool}. We now briefly review related work relying on both supervised and reinforcement learning.

    \noindent \textbf{Supervised learning:}
        Attempts have been made to train a classifier to predict whether a vertex in a graph can be removed from consideration without effecting the quality of the solution found by a heuristic. For instance, the classifier can be used to prune the graph in one attempt \cite{sun2019using, sun2021generalization} or in a multi-stage approach where the graph is repeatedly pruned \cite{grassia2019learning, lauri2020learning}. These methods, however, rely on hand crafted features that are typically specific to the problems they are trying to solve, whereas LeNSE works with multiple budget-constrained problems. Several other methods look to use datasets labelled by heuristics to learn solutions directly \cite{vinyals2015pointer, khalil2016learning, joshi2019efficient, gasse2019exact, liu2021learning}; in contrast, our graph pruning approach consists of several incremental steps. A non-parametric graph pruning method is described by \citet{manchanda2020gcomb}; they use a ranking rule learnt from solutions provided by a heuristic.
        
    \noindent \textbf{Reinforcement learning:} 
        Finding sub-optimal solutions to a CO problem can be formulated as a sequential decision problem, which is typically modelled as a Markov Decision Process. \citet{bello2016neural} was the first attempt to use a policy gradient algorithm and pointer networks \cite{vinyals2015pointer} to solve the Travelling Salesman Problem (TSP). Further improvements geared towards modelling the relational structure of the problem are found in \citet{dai2017learning}, which used a Q-function with a GNN \cite{dai2016discriminative}. The algorithm learns the value of adding a vertex at a time and uses these values to sequentially build a solution. A number of other sequential approaches have also appeared in the literature \cite{kool2018attention, ma2019combinatorial, li2019disco, almasan2019deep, bai2021glsearch}. In particular, the Ranked Reward algorithm \cite{laterre2018ranked} framed a CO problem as a single player game that can be solved through self-play. There has also been work that start with a solution and make perturbations to it until some termination criteria is met \cite{barrett2020exploratory, yao2021reversible}. Recently, \citet{wang2021bi} proposed an algorithm that learns to edit graphs by the removal/addition/modification of edges where the reward is driven by a heuristic deployed on the modified graph. The algorithm starts with the entire graph and looks to sequentially remove edges, requiring multiple forward passes of a GNN to embed the entire graph at each step of an episode; the computational overheads become prohibitive when using large graphs. In our work, we also use GNNs to learn graph embeddings; however, these embeddings only involve subgraphs that are significantly smaller than the original graph.
        
\section{Methodology}
    \subsection*{Problem formulation}
        
        We are given a CO problem defined over a graph $G = (V, E)$, where $V$ is the vertex set and $E$ is the edge set.   In this work we are concerned with budget-constrained problems, and rely on existing heuristic algorithms that can find near-optimal solutions. An optimal solution is defined as a subset of vertices $X^* \in \mathcal{X} = \{X: X \subset V, | X | = b\}$ that maximises the objective function $f(\cdot)$, i.e.
            \begin{equation}
                X^* = \underset{X \in \mathcal{X}}{\arg\max}\; f(X) \;;
            \end{equation}
            where $\mathcal{X}$ is the space of feasible solutions and $b$ is the available budget.  
            The problem we set out to address consists of finding an optimal subgraph of $G$ which contains the optimal $b$ vertices, but has much fewer vertices than $G$. More precisely, if we let $\mathcal{H}(\cdot)$ denote the heuristic solver taking a graph $G$ as input and returning an optimal solution $X^*$, we are interested in finding a subgraph $S=(V_S, E_S)$ containing $M$ vertices, with $b < M \mll  |V|$ and such that 
            \begin{equation}
                \label{eq: objective}
                f\left(\mathcal{H}(S)\right) / f\left( X^* \right) = 1\;.
            \end{equation}
           In practice, this often yields $|E_S| \mll  |E|$. 
            
        
    \subsection*{Learning a discriminative subgraph representation}
        \label{sec: subgraph representation}

        Our initial objective is to learn a Euclidean subgraph embedding for all subgraphs of $G$ with the required number of vertices, which will later be used as a \emph{navigation map} for LeNSE. For this purpose, we introduce an encoder $\psi : \mathcal{G} \rightarrow \mathbb{R}^d$ to map such subgraphs onto a $d$-dimensional space, where $\mathcal{G}$ is the set of subgraphs of $G$. An essential property we require is that  the coordinates of the embedded subgraph are informative about  the subgraph's likelihood of containing a nearly-optimal solution. 
        
        To achieve this, we frame the problem as one of subgraph classification, and we assume that every subgraph $S$ can be assigned a label from the set $\{1, 2, ..., K\}$. The labels correspond to rankings, where the highest rank indicates that $S$ is very likely to contain the best possible solution whilst lower ranks are associated with subgraphs expected to lead to worse solutions.
        
        In order to train the encoder, we use only a randomly selected and small portion of the entire graph, $G_T \subset G$, for which an optimal solution $X^*$ can be readily obtained. A training dataset is then generated by randomly sampling $N$ subgraphs of $G_T$ with the required size $M$, using Equation \eqref{eq: objective} as a proxy to determine their label (see Appendix \ref{sec: dataset generation}). This labelling mechanism ensures that only the relative quality of a subgraph compared to other ones is used to drive the learning process. This process is heuristic-agnostic hence can be flexibly deployed for other CO problems and/or heuristics. 
      
        To facilitate the process of learning a navigation policy, the encoder should learn a representation such that subgraphs sharing the same label are clustered together, forming approximately uni-modal point clouds, whilst also maintaining a strictly monotonic ordering across clusters. With these requirements in mind, we parameterise $\psi$ as a GNN that consists of convolutional and differentiable coarsening layers, whose weights are learned by minimizing the InfoNCE loss \cite{oord2018representation, chen2020simple, he2020momentum}:
            \begin{equation}
                \label{eq: contrastive loss}
                \mathcal{L}(S) = -\log \left( \frac{\exp(x \cdot x_+ / \tau)}{\exp(x \cdot x_+ / \tau) + \sum\limits_{i=0}^k \exp(x \cdot x_-^{(i)}/ \tau))} \right) 
            \end{equation}
        where $S$ is the input subgraph, $x = \psi(S)$ is its encoded representation, $x_+ = \psi(S_+)$ is the embedding of a subgraph sharing the same label as $S$ (positive sample), $\{x_-^{(0)}, ..., x_-^{(k)}\}$ is the set of subgraph embeddings (i.e. $x_-^{(i)} = \psi(S_-^{(i)}$)) for subgraphs belonging to different classes (negative samples), and $\tau$ is a temperature hyper-parameter. 
        
        The InfoNCE loss is a contrastive predictive coding (CPC) based function which maximises the mutual information between the query subgraph and a positive sample, whilst using negative samples as anchors to prevent a collapse of the embedding space. We have found that minimising the InfoNCE, in comparison to other losses such as ordinal and cross entropy, provides a good trade-off between maintaining the geometric structure of the embedded space that we require whilst also achieving good classification performance (see also Appendix \ref{sec: loss ablation study} for an ablation study). CPC-based approaches have also been shown to have superior performance in downstream tasks compared to other losses \cite{song2020multi}.

    \subsection*{Subgraph navigation as a Markov Decision Process}
        
        The process of incrementally exploring subgraphs, starting from a randomly chosen one, is delegated to an agent that follows an optimal policy. To learn such a policy, we consider the traditional Markov Decision Process (MDP) setting \cite{bellman1957markovian}. We are given a graph $G$ and a subgraph mapping function $\xi : \Lambda \rightarrow \mathcal{G}$, where $\Lambda$ is the power set of $V$ and $\mathcal{G}$ is the set of subgraphs of $G$. In principle $\xi(\cdot)$ can be any function which returns a subgraph given a set of nodes, but we give the precise definition used in this work in the subsequent section. The MDP is defined by a tuple $(\mathcal{S}, \mathcal{A}, \mathcal{T}, \mathcal{R}, \rho_0, \gamma)$ whose elements are introduced below. 
        
        The state space, $\mathcal{S}$, consists of the set of subgraphs induced by $\xi$ for vertex sets of fixed size $M \in \mathbb{N}$, i.e. $\mathcal{S} = \{ \xi(X) : X \in \Lambda, |X| = M\} \subset \mathcal{G}$; we say that $s_t = \xi(X_t)$ is the state(/subgraph) induced by selected vertices $X_t$. The action space includes a subgraph updating operation; taking an action $a_t \in \mathcal{A}$ modifies the current subgraph $s_t \in \mathcal{S}$ into a new subgraph $s_{t+1} \in \mathcal{S}$. There is flexibility in how this modification operation can be defined, and in the subsequent section we discuss the specific operation used in this work. The transition function $\mathcal{T} : \mathcal{S} \times \mathcal{A} \rightarrow \mathcal{S}$ is a deterministic function that, given the current subgraph and action, returns the next subgraph. Specifically, to obtain $s_{t+1}$ from $s_t$, the agent first modifies $X_t$ into $X_{t+1}$, and then the transition function uses the subgraph mapping function $\xi$ to obtain the subgraph $s_{t+1} = \xi(X_{t+1})$.
        
        The reward function $\mathcal{R} : \mathcal{S} \times \mathcal{A} \rightarrow \mathbb{R}$ measures the distance between the current state (i.e. subgraph) and the region containing the highest quality subgraphs, and is evaluated using the embedding space, as follows. We let $\{S^{(1)}, ... , S^{(n)}\}$ be the set of subgraphs belonging to class 1 in the training dataset, and define $g^*$ to be the centroid of $\{\psi(S^{(i)})\}_{i = 1}^n$. Throughout we will refer to $g^*$ as the goal. The reward is then defined as
            \begin{equation}
                \label{eq: mdp reward}
                \mathcal{R}(s_t, a_t) = r_{t+1} = -\beta \times \| g^* - s_{t+1} \|_2 \; ;
            \end{equation}
        where $\beta \in (0, \infty)$ is a scaling parameter. The desired policy $\pi : \mathcal{S} \rightarrow [0, 1]$ maps a state on to a state-conditioned distribution over actions and maximises the expected (discounted) rewards 
        \begin{equation}
            \mathbb{E}_{\tau \sim \rho_0, \pi}\left[\sum_{t=0}^\infty \gamma^t r_{t+1}\right] \; ;
        \end{equation}
        where $\tau = (s_0, a_1, ..., a_{T-1}, s_T)$ is the trajectory generated by the initial state distribution $\rho_0$ and the policy $\pi$, and $\gamma \in [0,1)$ is a discount factor. This means that, in relation to the navigation task, the agent should move incrementally closer to the goal when following a trained policy. Figure \ref{fig: traceplot example} shows a real trajectory followed by the agent in a $2$-dimensional embedding space (A) and the corresponding distance from the goal averaged over ten episodes (B).

    \subsection*{Subgraph updating operation}
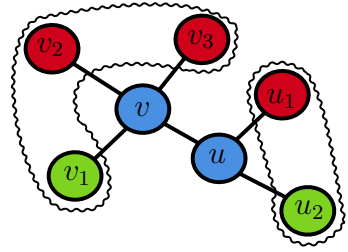
\begin{figure}[tp]
                \centering
                \tikzset{every picture/.style={line width=0.75pt}} 
        
                \begin{tikzpicture}[x=0.4pt,y=0.4pt,yscale=-0.9,xscale=1]
                
                \draw  [fill={rgb, 255:red, 74; green, 144; blue, 226 }  ,fill opacity=1 ][line width=1.5]  (132.26,109.96) .. controls (140.63,98.98) and (156.32,96.86) .. (167.3,105.23) .. controls (178.28,113.59) and (180.4,129.28) .. (172.04,140.26) .. controls (163.67,151.25) and (147.98,153.37) .. (137,145) .. controls (126.02,136.63) and (123.9,120.95) .. (132.26,109.96) -- cycle ;
                \draw  [fill={rgb, 255:red, 208; green, 2; blue, 27 }  ,fill opacity=1 ][line width=1.5]  (46.55,46.29) .. controls (55.03,35.39) and (70.74,33.43) .. (81.63,41.92) .. controls (92.53,50.4) and (94.48,66.11) .. (86,77) .. controls (77.52,87.89) and (61.81,89.85) .. (50.92,81.37) .. controls (40.02,72.89) and (38.06,57.18) .. (46.55,46.29) -- cycle ;
                \draw  [fill={rgb, 255:red, 126; green, 211; blue, 33 }  ,fill opacity=1 ][line width=1.5]  (70.65,177.49) .. controls (80.55,167.87) and (96.38,168.1) .. (106,178) .. controls (115.62,187.9) and (115.39,203.73) .. (105.49,213.35) .. controls (95.58,222.97) and (79.76,222.74) .. (70.14,212.84) .. controls (60.52,202.94) and (60.74,187.11) .. (70.65,177.49) -- cycle ;
                \draw  [fill={rgb, 255:red, 208; green, 2; blue, 27 }  ,fill opacity=1 ][line width=1.5]  (188.82,35.79) .. controls (197.67,25.19) and (213.43,23.76) .. (224.03,32.61) .. controls (234.63,41.45) and (236.06,57.22) .. (227.21,67.82) .. controls (218.37,78.42) and (202.6,79.85) .. (192,71) .. controls (181.4,62.15) and (179.97,46.39) .. (188.82,35.79) -- cycle ;
                \draw [line width=1.5]    (86,77) -- (132.26,109.96) ;
                \draw [line width=1.5]    (106,178) -- (137,145) ;
                \draw [line width=1.5]    (167.3,105.23) -- (192,71) ;
                \draw  [fill={rgb, 255:red, 208; green, 2; blue, 27 }  ,fill opacity=1 ][line width=1.5]  (267.57,128.84) .. controls (257.16,119.76) and (256.08,103.97) .. (265.16,93.57) .. controls (274.24,83.16) and (290.03,82.08) .. (300.43,91.16) .. controls (310.84,100.24) and (311.92,116.03) .. (302.84,126.43) .. controls (293.76,136.84) and (277.97,137.92) .. (267.57,128.84) -- cycle ;
                \draw  [fill={rgb, 255:red, 126; green, 211; blue, 33 }  ,fill opacity=1 ][line width=1.5]  (287.53,217.98) .. controls (295.34,206.59) and (310.9,203.69) .. (322.28,211.49) .. controls (333.67,219.3) and (336.58,234.86) .. (328.77,246.25) .. controls (320.96,257.64) and (305.4,260.54) .. (294.01,252.73) .. controls (282.63,244.93) and (279.72,229.37) .. (287.53,217.98) -- cycle ;
                \draw  [fill={rgb, 255:red, 74; green, 144; blue, 226 }  ,fill opacity=1 ][line width=1.5]  (205,161) .. controls (213.79,150.35) and (229.55,148.85) .. (240.2,157.64) .. controls (250.84,166.43) and (252.35,182.19) .. (243.55,192.84) .. controls (234.76,203.48) and (219,204.99) .. (208.36,196.2) .. controls (197.71,187.4) and (196.21,171.65) .. (205,161) -- cycle ;
                \draw [line width=1.5]    (172.04,140.26) -- (205,161) ;
                \draw [line width=1.5]    (243.55,192.84) -- (287.53,217.98) ;
                \draw [line width=1.5]    (267.57,128.84) -- (240.2,157.64) ;
                \draw    (87,231) .. controls (85.24,232.99) and (83.58,233.17) .. (82.01,231.55) .. controls (80.46,229.71) and (78.61,229.37) .. (76.46,230.52) .. controls (74.41,231.49) and (73.07,230.88) .. (72.42,228.67) .. controls (71.65,226.18) and (70.12,225.1) .. (67.85,225.43) .. controls (65.42,225.48) and (64.36,224.48) .. (64.68,222.44) .. controls (64.71,219.99) and (63.49,218.57) .. (61.01,218.17) .. controls (58.87,218.1) and (57.89,216.74) .. (58.07,214.08) .. controls (58.71,212.01) and (57.95,210.82) .. (55.79,210.49) .. controls (53.22,209.39) and (52.31,207.77) .. (53.04,205.62) .. controls (53.81,203.5) and (53.11,202.11) .. (50.92,201.45) .. controls (48.73,200.7) and (48.05,199.23) .. (48.89,197.05) .. controls (49.76,194.88) and (49.11,193.34) .. (46.94,192.43) .. controls (44.75,191.44) and (44.13,189.84) .. (45.07,187.61) .. controls (46.04,185.4) and (45.45,183.74) .. (43.29,182.62) .. controls (41.14,181.43) and (40.58,179.71) .. (41.61,177.46) .. controls (42.67,175.23) and (42.27,173.91) .. (40.41,173.5) .. controls (38.29,172.16) and (37.78,170.36) .. (38.89,168.09) .. controls (40.02,165.85) and (39.67,164.48) .. (37.82,163.97) .. controls (35.75,162.5) and (35.3,160.63) .. (36.49,158.37) .. controls (37.7,156.13) and (37.38,154.71) .. (35.55,154.12) .. controls (33.53,152.54) and (33.15,150.63) .. (34.41,148.39) .. controls (35.87,147.12) and (35.61,145.68) .. (33.62,144.05) .. controls (31.65,142.38) and (31.33,140.44) .. (32.67,138.23) .. controls (34.18,137.02) and (33.97,135.56) .. (32.04,133.85) .. controls (30.13,132.1) and (29.94,130.64) .. (31.47,129.45) .. controls (32.9,127.32) and (32.69,125.37) .. (30.82,123.6) .. controls (29.07,122.78) and (28.93,121.32) .. (30.42,119.22) .. controls (31.93,117.15) and (31.79,115.21) .. (30,113.4) .. controls (28.27,112.53) and (28.19,111.08) .. (29.76,109.07) .. controls (31.35,107.1) and (31.29,105.19) .. (29.57,103.34) .. controls (27.89,102.42) and (27.87,101) .. (29.52,99.09) .. controls (31.19,97.24) and (31.2,95.38) .. (29.57,93.51) .. controls (27.92,92.54) and (27.97,91.17) .. (29.7,89.39) .. controls (31.45,87.68) and (31.55,85.88) .. (30,84) .. controls (28.48,82.09) and (28.63,80.34) .. (30.45,78.75) .. controls (32.3,77.23) and (32.5,75.53) .. (31.06,73.65) .. controls (29.65,71.74) and (29.9,70.09) .. (31.81,68.71) .. controls (33.74,67.43) and (34.05,65.85) .. (32.73,63.96) .. controls (31.46,62.05) and (31.82,60.53) .. (33.81,59.42) .. controls (35.81,58.42) and (36.22,56.98) .. (35.05,55.09) .. controls (34.18,52.46) and (34.77,50.77) .. (36.84,50.02) .. controls (38.9,49.4) and (39.59,47.84) .. (38.9,45.35) .. controls (38.03,43.4) and (38.81,41.99) .. (41.23,41.12) .. controls (43.61,40.54) and (44.48,39.28) .. (43.84,37.35) .. controls (43.75,34.88) and (44.92,33.6) .. (47.36,33.5) .. controls (49.66,33.73) and (50.97,32.7) .. (51.3,30.43) -- (52,30) ;
                \draw    (52,30) .. controls (53.61,27.9) and (55.27,27.5) .. (57,28.81) .. controls (59.4,29.96) and (61.09,29.57) .. (62.08,27.64) .. controls (63.79,25.55) and (65.51,25.17) .. (67.25,26.5) .. controls (68.99,27.83) and (70.74,27.46) .. (72.49,25.39) .. controls (73.57,23.47) and (74.98,23.18) .. (76.73,24.53) .. controls (78.48,25.88) and (80.26,25.53) .. (82.08,23.48) .. controls (83.21,21.57) and (84.65,21.31) .. (86.4,22.68) .. controls (88.87,23.93) and (90.68,23.6) .. (91.84,21.71) .. controls (93.02,19.82) and (94.48,19.57) .. (96.21,20.97) .. controls (98.67,22.26) and (100.5,21.97) .. (101.71,20.09) .. controls (102.93,18.22) and (104.4,18) .. (106.12,19.43) .. controls (108.56,20.77) and (110.4,20.51) .. (111.65,18.66) .. controls (112.91,16.81) and (114.38,16.62) .. (116.07,18.08) .. controls (118.48,19.47) and (120.32,19.25) .. (121.6,17.43) .. controls (122.89,15.61) and (124.37,15.45) .. (126.02,16.95) .. controls (128.39,18.39) and (130.22,18.21) .. (131.53,16.42) .. controls (132.86,14.63) and (134.32,14.51) .. (135.92,16.05) .. controls (138.23,17.55) and (140.05,17.42) .. (141.39,15.67) .. controls (142.74,13.92) and (144.55,13.81) .. (146.81,15.36) .. controls (148.32,16.96) and (149.75,16.9) .. (151.12,15.18) .. controls (152.5,13.47) and (154.28,13.43) .. (156.45,15.04) .. controls (157.88,16.69) and (159.64,16.67) .. (161.73,14.99) .. controls (163.14,13.33) and (164.53,13.34) .. (165.9,15.03) .. controls (167.91,16.74) and (169.63,16.78) .. (171.04,15.16) .. controls (173.15,13.58) and (174.84,13.66) .. (176.1,15.4) .. controls (177.98,17.19) and (179.64,17.31) .. (181.07,15.75) .. controls (183.19,14.27) and (184.81,14.43) .. (185.94,16.22) .. controls (187.65,18.09) and (189.24,18.29) .. (190.71,16.8) .. controls (192.83,15.42) and (194.68,15.71) .. (196.27,17.66) .. controls (197.17,19.51) and (198.67,19.8) .. (200.76,18.51) .. controls (202.89,17.27) and (204.63,17.67) .. (205.98,19.71) .. controls (207.22,21.76) and (208.89,22.23) .. (210.99,21.11) .. controls (212.58,19.86) and (213.92,20.3) .. (214.99,22.42) .. controls (215.94,24.55) and (217.72,25.24) .. (220.33,24.51) .. controls (222.49,23.66) and (223.92,24.33) .. (224.63,26.53) .. controls (225.19,28.72) and (226.54,29.46) .. (228.67,28.77) .. controls (231.3,28.47) and (232.75,29.44) .. (233.03,31.68) .. controls (233.15,33.87) and (234.47,34.95) .. (236.99,34.9) .. controls (239.21,34.71) and (240.39,35.89) .. (240.53,38.46) -- (241,39) ;
                \draw    (87,231) .. controls (88.89,229.54) and (90.58,229.7) .. (92.07,231.48) .. controls (94.16,233.19) and (95.93,233.18) .. (97.38,231.46) .. controls (98.57,229.68) and (100.14,229.48) .. (102.07,230.85) .. controls (104.15,232.06) and (105.87,231.53) .. (107.24,229.27) .. controls (107.76,227.16) and (109.17,226.36) .. (111.47,226.87) .. controls (113.66,227.23) and (114.79,226.18) .. (114.84,223.73) .. controls (114.46,221.46) and (115.39,220.04) .. (117.63,219.47) .. controls (119.88,218.46) and (120.5,216.82) .. (119.49,214.53) .. controls (118.19,212.78) and (118.51,211.13) .. (120.44,209.58) .. controls (122.24,208.14) and (122.34,206.57) .. (120.74,204.86) .. controls (119.06,203.28) and (119.01,201.63) .. (120.58,199.9) .. controls (122.08,198.01) and (121.89,196.29) .. (120,194.75) .. controls (118.07,193.3) and (117.79,191.77) .. (119.17,190.14) .. controls (120.4,187.98) and (119.98,186.2) .. (117.91,184.81) .. controls (115.94,183.93) and (115.48,182.37) .. (116.54,180.13) .. controls (117.69,178.26) and (117.16,176.71) .. (114.96,175.48) .. controls (112.75,174.33) and (112.16,172.8) .. (113.2,170.89) .. controls (114.01,168.52) and (113.36,167.03) .. (111.27,166.42) .. controls (109,165.48) and (108.21,163.84) .. (108.92,161.49) .. controls (109.78,159.52) and (109.05,158.15) .. (106.74,157.38) .. controls (104.45,156.73) and (103.59,155.26) .. (104.16,152.97) .. controls (104.68,150.68) and (103.8,149.33) .. (101.52,148.93) .. controls (99.08,148.38) and (98.08,147.03) .. (98.53,144.88) .. controls (98.68,142.46) and (97.59,141.21) .. (95.24,141.12) -- (93,139) ;
                \draw    (93,139) .. controls (90.61,137.3) and (89.97,135.45) .. (91.07,133.46) .. controls (92.24,131.59) and (91.84,130.19) .. (89.86,129.27) .. controls (87.75,127.61) and (87.38,125.95) .. (88.75,124.3) .. controls (90.1,122.19) and (89.88,120.64) .. (88.08,119.65) .. controls (86.29,117.96) and (86.2,116.23) .. (87.83,114.45) .. controls (89.55,112.93) and (89.65,111.34) .. (88.14,109.67) .. controls (86.85,107.34) and (87.2,105.64) .. (89.18,104.59) .. controls (91.33,103.42) and (91.91,101.91) .. (90.9,100.04) .. controls (90.31,97.61) and (91.22,96.07) .. (93.63,95.44) .. controls (95.98,95.16) and (96.98,93.99) .. (96.63,91.93) .. controls (96.84,89.46) and (98.19,88.3) .. (100.68,88.46) .. controls (103.07,88.85) and (104.42,87.99) .. (104.72,85.88) .. controls (105.67,83.51) and (107.15,82.79) .. (109.17,83.7) .. controls (111.55,84.54) and (113.15,83.94) .. (113.98,81.89) .. controls (114.95,79.84) and (116.66,79.36) .. (119.09,80.43) .. controls (120.95,81.72) and (122.48,81.39) .. (123.67,79.45) .. controls (124.98,77.53) and (126.56,77.28) .. (128.41,78.69) .. controls (130.2,80.15) and (132.1,79.95) .. (134.09,78.08) .. controls (135.63,76.29) and (137.28,76.19) .. (139.05,77.76) .. controls (140.76,79.37) and (142.43,79.32) .. (144.06,77.63) .. controls (145.75,75.96) and (147.42,75.97) .. (149.09,77.66) .. controls (150.7,79.37) and (152.1,79.42) .. (153.28,77.8) .. controls (155.05,76.23) and (156.71,76.33) .. (158.26,78.11) .. controls (160.3,79.95) and (162.21,80.12) .. (163.98,78.61) .. controls (165.78,77.13) and (167.38,77.32) .. (168.77,79.17) .. controls (170.1,81.03) and (171.65,81.24) .. (173.42,79.81) .. controls (175.19,78.4) and (176.92,78.69) .. (178.61,80.67) .. controls (179.73,82.57) and (181.37,82.88) .. (183.53,81.61) .. controls (185.22,80.26) and (186.74,80.6) .. (188.1,82.61) .. controls (189.32,84.62) and (190.9,85.01) .. (192.84,83.8) -- (197,85) ;
                \draw    (197,85) .. controls (198.88,83.56) and (200.73,83.77) .. (202.55,85.63) .. controls (203.99,87.4) and (205.49,87.49) .. (207.06,85.9) .. controls (208.8,84.25) and (210.5,84.24) .. (212.17,85.85) .. controls (213.9,87.4) and (215.56,87.23) .. (217.15,85.35) .. controls (218.42,83.41) and (219.99,83.05) .. (221.88,84.28) .. controls (224.13,85.25) and (225.67,84.63) .. (226.51,82.4) .. controls (227.08,80.11) and (228.5,79.13) .. (230.78,79.46) .. controls (233.19,79.4) and (234.29,78.17) .. (234.06,75.78) .. controls (233.49,73.57) and (234.3,72.11) .. (236.51,71.42) .. controls (238.72,70.37) and (239.27,68.83) .. (238.17,66.8) .. controls (236.94,64.88) and (237.32,63.19) .. (239.31,61.72) .. controls (241.18,60.53) and (241.41,58.95) .. (239.98,57) .. controls (238.49,55.09) and (238.64,53.43) .. (240.42,52.03) .. controls (242.19,50.28) and (242.29,48.55) .. (240.7,46.85) .. controls (239.11,45.16) and (239.17,43.51) .. (240.9,41.92) -- (241,39) ;
                \draw    (277,213) .. controls (278.25,215.17) and (277.9,217.03) .. (275.95,218.56) .. controls (274.12,219.75) and (273.96,221.32) .. (275.47,223.29) .. controls (277.07,225.08) and (277.04,226.61) .. (275.38,227.88) .. controls (273.79,229.69) and (273.9,231.37) .. (275.71,232.92) .. controls (277.58,234.27) and (277.89,236.07) .. (276.63,238.3) .. controls (275.4,240.27) and (275.82,241.78) .. (277.88,242.81) .. controls (280.08,243.96) and (280.71,245.54) .. (279.76,247.53) .. controls (279.12,249.96) and (279.89,251.39) .. (282.08,251.84) .. controls (284.41,252.35) and (285.43,253.77) .. (285.14,256.12) .. controls (284.82,258.28) and (285.87,259.4) .. (288.28,259.47) .. controls (290.58,259.26) and (291.87,260.3) .. (292.14,262.58) .. controls (292.99,265.07) and (294.53,265.95) .. (296.78,265.22) .. controls (298.83,264.25) and (300.33,264.79) .. (301.3,266.84) .. controls (302.57,268.83) and (304.31,269.12) .. (306.52,267.72) .. controls (308.12,266.09) and (309.76,266.06) .. (311.44,267.63) .. controls (313.42,269.02) and (315.09,268.7) .. (316.46,266.66) .. controls (317.25,264.66) and (318.77,264.1) .. (321.02,264.99) .. controls (323.15,265.8) and (324.68,264.97) .. (325.59,262.52) .. controls (325.96,260.26) and (327.3,259.3) .. (329.61,259.63) .. controls (332.05,259.74) and (333.2,258.71) .. (333.07,256.55) .. controls (333.15,254.08) and (334.29,252.88) .. (336.48,252.93) .. controls (339.06,252.41) and (340.16,251.02) .. (339.79,248.75) .. controls (339.34,246.5) and (340.26,245.15) .. (342.55,244.71) -- (343,244) ;
                \draw    (343,244) .. controls (341.06,242.63) and (340.76,240.97) .. (342.11,239) .. controls (343.46,237.03) and (343.16,235.33) .. (341.23,233.91) .. controls (339.3,232.47) and (339.01,230.75) .. (340.36,228.74) .. controls (341.72,226.73) and (341.43,224.98) .. (339.5,223.49) .. controls (337.67,222.58) and (337.43,221.11) .. (338.79,219.07) .. controls (340.16,217.04) and (339.88,215.26) .. (337.95,213.73) .. controls (336.11,212.79) and (335.87,211.29) .. (337.24,209.24) .. controls (338.61,207.18) and (338.32,205.38) .. (336.39,203.83) .. controls (334.56,202.88) and (334.32,201.37) .. (335.69,199.3) .. controls (337.05,197.23) and (336.76,195.41) .. (334.83,193.85) .. controls (333,192.9) and (332.76,191.38) .. (334.12,189.31) .. controls (335.47,187.23) and (335.23,185.72) .. (333.4,184.77) .. controls (331.46,183.22) and (331.17,181.41) .. (332.52,179.34) .. controls (333.87,177.27) and (333.62,175.77) .. (331.78,174.84) .. controls (329.83,173.32) and (329.53,171.53) .. (330.88,169.46) .. controls (332.21,167.39) and (331.96,165.91) .. (330.11,165.02) .. controls (328.16,163.54) and (327.84,161.78) .. (329.17,159.73) .. controls (330.49,157.68) and (330.17,155.94) .. (328.21,154.51) .. controls (326.25,153.1) and (325.98,151.67) .. (327.39,150.22) .. controls (328.7,148.19) and (328.36,146.5) .. (326.39,145.14) .. controls (324.42,143.81) and (324.07,142.15) .. (325.35,140.16) .. controls (326.62,138.17) and (326.27,136.54) .. (324.28,135.28) .. controls (322.29,134.06) and (321.93,132.47) .. (323.18,130.52) .. controls (324.43,128.57) and (323.98,126.77) .. (321.85,125.12) .. controls (319.85,124.04) and (319.46,122.55) .. (320.67,120.64) .. controls (321.87,118.73) and (321.39,117.06) .. (319.23,115.61) .. controls (317.22,114.7) and (316.72,113.1) .. (317.74,110.79) .. controls (318.88,108.93) and (318.36,107.4) .. (316.18,106.21) .. controls (314.16,105.54) and (313.62,104.09) .. (314.55,101.88) .. controls (315.45,99.67) and (314.8,98.13) .. (312.61,97.27) .. controls (310.43,96.54) and (309.66,94.96) .. (310.29,92.52) .. controls (311.02,90.4) and (310.21,88.99) .. (307.85,88.29) -- (307,87) ;
                \draw    (253,120) .. controls (251.05,118.35) and (250.81,116.68) .. (252.29,114.99) .. controls (253.83,113.12) and (253.75,111.36) .. (252.05,109.71) .. controls (250.44,107.93) and (250.53,106.27) .. (252.34,104.74) .. controls (254.21,103.45) and (254.47,101.91) .. (253.13,100.12) .. controls (252.01,97.89) and (252.53,96.19) .. (254.7,95.03) .. controls (256.88,94.2) and (257.61,92.69) .. (256.89,90.5) .. controls (256.38,88.18) and (257.3,86.86) .. (259.63,86.53) .. controls (261.87,86.52) and (263.05,85.31) .. (263.16,82.92) .. controls (263.39,80.62) and (264.74,79.67) .. (267.2,80.06) .. controls (269.26,80.86) and (270.75,80.17) .. (271.68,77.98) .. controls (272.99,75.85) and (274.61,75.43) .. (276.53,76.74) .. controls (278.22,78.21) and (279.94,78.09) .. (281.67,76.36) .. controls (283.45,74.79) and (285.09,74.94) .. (286.6,76.82) .. controls (287.87,78.79) and (289.42,79.16) .. (291.23,77.94) .. controls (293.55,76.99) and (295.12,77.6) .. (295.93,79.77) .. controls (296.57,81.95) and (298.14,82.8) .. (300.63,82.32) .. controls (302.71,81.68) and (303.99,82.56) .. (304.47,84.96) -- (307,87) ;
                \draw    (253,120) .. controls (255.03,121.19) and (255.45,122.81) .. (254.25,124.84) .. controls (253.05,126.87) and (253.47,128.49) .. (255.5,129.68) .. controls (257.53,130.87) and (257.95,132.49) .. (256.75,134.52) .. controls (255.55,136.55) and (255.97,138.17) .. (258,139.37) .. controls (260.03,140.56) and (260.45,142.18) .. (259.25,144.21) .. controls (258.05,146.24) and (258.47,147.86) .. (260.5,149.05) .. controls (262.53,150.24) and (262.95,151.86) .. (261.75,153.89) .. controls (260.55,155.92) and (260.97,157.54) .. (263,158.73) .. controls (265.03,159.93) and (265.44,161.54) .. (264.24,163.57) .. controls (263.04,165.6) and (263.46,167.22) .. (265.49,168.41) .. controls (267.52,169.61) and (267.94,171.23) .. (266.74,173.26) .. controls (265.54,175.29) and (265.96,176.91) .. (267.99,178.1) .. controls (270.02,179.29) and (270.44,180.91) .. (269.24,182.94) .. controls (268.04,184.97) and (268.46,186.59) .. (270.49,187.78) .. controls (272.52,188.97) and (272.94,190.59) .. (271.74,192.62) .. controls (270.54,194.65) and (270.96,196.27) .. (272.99,197.46) .. controls (275.02,198.65) and (275.44,200.27) .. (274.24,202.3) .. controls (273.04,204.33) and (273.46,205.95) .. (275.49,207.14) .. controls (277.52,208.34) and (277.94,209.96) .. (276.74,211.99) -- (277,213) -- (277,213) ;
                
                \draw (152,125) node[inner sep=0.75pt]  [font=\large]  {$v$};
                \draw (90, 195) node [inner sep=0.75pt]    [font=\large]{$v_{1}$};
                \draw (65, 60) node [inner sep=0.75pt]    [font=\large]{$v_{2}$};
                \draw (207.5,51.5) node [inner sep=0.75pt]    [font=\large]{$v_{3}$};
                \draw (222.5,175) node [inner sep=0.75pt]    [font=\large]{$u$};
                \draw (283.5, 114) node [inner sep=0.75pt]    [font=\large]{$u_{1}$};
                \draw (310,231) node [inner sep=0.75pt]    [font=\large]{$u_{2}$};

                \end{tikzpicture}

                \caption{Illustration of the subgraph modification operation. The current state is given by $\xi(X_t)$, where $X_t = \{u, v\}$ (blue vertices). For each vertex in $X_t$, a neighbour is sampled to form the action tuples. The chosen vertices are highlighted in green whilst the vertices considered, but not chosen, are highlighted in red. For instance, for vertex $v$, $v_1$ is sampled from its one-hop neighbours, $\{v_1, v_2, v_3\}$, uniformly at random. The action space is thus $\mathcal{A}_t = \{a(v, v_1), a(u, u_2)\}$.}
                \label{fig: action selection}
            \end{figure}

        In the design of the subgraph updating operation, we faced the challenge of keeping the action space small whilst being able to change the connectivity structure of the current subgraph. Our proposed solution is to allow the agent's action to update $X_{t}$ by replacing a single vertex at a time. For each vertex $v \in X_t$, a candidate vertex $u$ is randomly sampled from the one-hop neighbourhood of $v$, and the agent can then decide to swap $v$ with $u$; this is indicated by $a(v, u)$. Since $|X_t| = M$, there are $M$ possible actions; see Figure \ref{fig: action selection}. 
        Once $X_{t+1}$ is obtained, the new subgraph $S = (V_S, E_S) \in \mathcal{G}$ is determined by $\xi(\cdot)$, as follows: 
        \begin{align}
            \label{eq: subgraph definition}
            \begin{split}
            V_S & = \bigcup_{v \in X_{t+1} } \left(\{v\} \cup \mathcal{N}(v)\right) \; ; \\
            E_S & = \{(u, v) \in E : u \in X_{t+1} \vee v \in X_{t+1} \} \; .
            \end{split}
        \end{align}
        As can be seen here, despite replacing only a single vertex in $X_t$ per time step, a substantial topological change can be achieved due to the change of neighbourhood of the updated vertex set. This allows LeNSE to learn how to move around the embedded space efficiently. 
        
        
    \subsection*{Q-Learning with guided exploration}
    
        To learn a subgraph navigation policy, we use Q-learning \cite{watkins1992q}. Specifically, given the complexity of the problem, we implement a Deep Q-Network (DQN) algorithm \cite{mnih2015human} whose network weights are parameterised by $\theta$. Typically, an $\epsilon$-greedy exploration policy is employed in Q-learning, where the policy takes a greedy action with probability $1 - \epsilon$ and acts randomly with probability $\epsilon$. As we are training LeNSE on a small portion of the graph, we can take advantage of the fact that, during the initial phase of learning a subgraph embedding, we run the heuristic on the entire training graph, hence we know what the optimal $b$ vertices are \emph{a-priori}. We propose to inject this prior knowledge into the exploration strategy by forcing the agent to select one of these optimal vertices when possible, thus $\emph{guiding}$ the agent towards the goal.
        
        If we let $B$ be the set of $b$ solution vertices, then when the vertex tuples $\{a(v_i, u_i)\}_{i=1}^M$ are sampled to add to the action space, we can check to see if any of the vertices $u_i$ are in $B$. If $u_i$ is in $B$ then the action tuple $a(v_i, u_i)$ can additionally be added to a set $\mathcal{B}_t$. If $\mathcal{B}_t$ is non-empty, instead of choosing a completely random action from $\mathcal{A}_t$, the agent randomly selects an action from $\mathcal{B}_t$. This will add one of the $b$ solution vertices to the subgraph and will move the agent in the right direction towards the goal $g^*$. Under the proposed guided exploration policy, an action is selected according to
        \begin{equation}
            a_t = 
                \begin{cases}
                    \mathcal{U}(\mathcal{A}_t) \;, & \mbox{ w.p. } \epsilon \times (1-\alpha) \; ; \\
                    \mathcal{U}(\mathcal{B}_t) \;, & \mbox{ w.p. } \epsilon \times \alpha \; ; \\
                    \underset{a \in \mathcal{A}_t}{\arg\max} \;\; Q(s_t, a; \theta)\;, & \mbox{ otherwise} \; .
                \end{cases}
        \end{equation}
        The parameter $\alpha$ controls the level of guidance and ensure that there is still sufficient exploration of the environment. Setting $\alpha = 0$ results in the standard $\epsilon$-greedy exploration policy. Note that this exploration policy can only be used at training time, as at test time it is assumed that the optimal $b$ vertices are unknown. The full LeNSE algorithm with guided exploration is detailed in Algorithm \ref{alg: LeNSE}.
        
        \begin{algorithm}[t!]
            \caption{LeNSE with Guided Exploration \label{alg: LeNSE}}
            \begin{algorithmic}[1]
                \REQUIRE Train graph $G_T = (V_T, E_T)$, Subgraph mapping function $\xi(\cdot)$, Subgraph size $M$, Set of $b$ solution vertices $B$, Replay memory $\mathcal{M} = \emptyset$, Number of episodes $N'$, Episode length $T$, Initial network parameters $\theta$, Update frequency $c$
                \FOR{episode in $1, 2, ..., N'$}
                    \STATE $X_0 \sim \mathcal{U}(V)$ such that $|X_0| = M$
                    \STATE Receive initial state $s_0 = \xi(X_0)$
                    \FOR{each $t$ in $1, ..., T$}
                        \STATE $\mathcal{A}_t = \emptyset$, $\mathcal{B}_t = \emptyset$
                        \FOR{$v \in X_t$}
                            \STATE $u \sim \mathcal{U}(\mathcal{N}(v) \setminus X_t)$
                            \STATE $\mathcal{A}_t = \mathcal{A}_t \cup \{a(v, u)\}$
                            \IF{$u$ in $B$}
                                \STATE $\mathcal{B}_t = \mathcal{B}_t \cup \{a(v, u)\}$
                            \ENDIF
                        \ENDFOR
                        \STATE Choose action $a_t$ according to guided exploration policy, receive reward $r_{t}$, update set of selected vertices $X_{t+1}$ and observe new state $s_{t+1} = \xi(X_{t+1})$
                        \STATE Add tuple $(s_t, a_t, r_{t}, s_{t+1})$ to $\mathcal{M}$
                        \IF{$t \equiv 0 \mbox{ mod } c$}
                            \STATE Sample random batch $R \sim \mathcal{U}(\mathcal{M})$ and update $\theta$ by SGD for $R$
                        \ENDIF
                    \ENDFOR
                \ENDFOR
            \end{algorithmic}
        \end{algorithm}

            \begin{figure*}[t]
                \centering
                    \includegraphics[width=\linewidth]{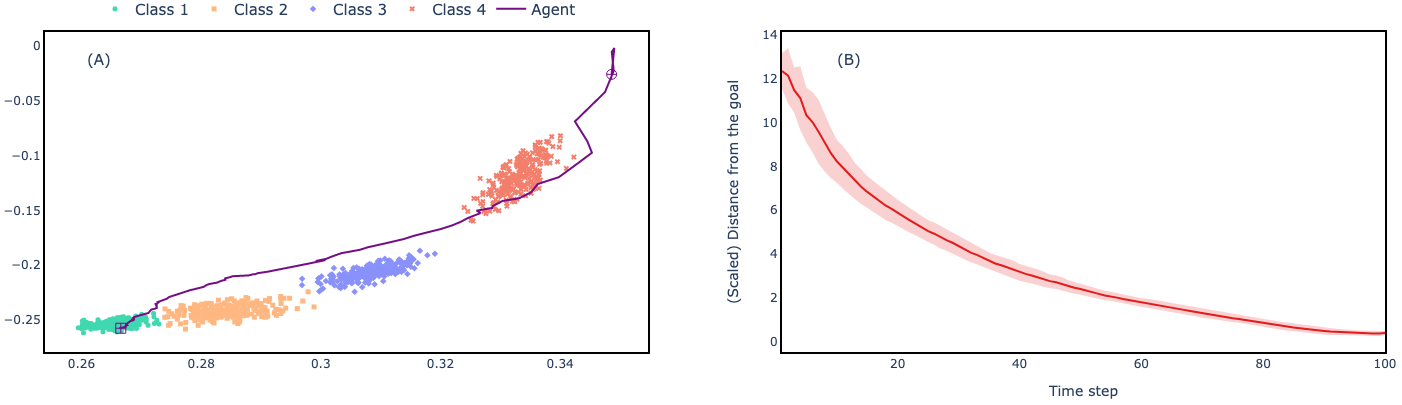}
                \caption{(A): An example of trajectory realised by the agent taken throughout a test episode. The goal for the agent is to reach the green region, i.e. the region of optimal subgraphs. The purple circle denotes the agents start position whilst the purple square denotes the agents final position. We used an autoencoder to map from the original embedding space to the 2-dimensional space in the Figure. (B): Time-series plots of how the distance from the optimal region (green in the plot on the left) changes over time. The red line is the mean distance (of 10 random episodes) per episode time step, with the shaded area corresponding to the 95\% confidence interval. Here the distance is computed in the original space (i.e. without using the autoencoder for dimensionality reduction) and is scaled using $\beta$ from Equation \eqref{eq: mdp reward}}.
                \label{fig: traceplot example}
            \end{figure*}
       
\section{Experimental studies} 
            
    \label{sec: experimental details}
        
    \subsection*{Setup}
        
    We extensively test the performance of LeNSE on the following three budget constrained problems:
            
        \noindent\textbf{Max Vertex Cover (MVC):}
            Given a graph $G = (V, E)$ and a budget $b$, find a set $X^*$ of $b$ vertices such that the coverage of $X^*$ is maximised, i.e. 
                \begin{equation}
                    X^* = \underset{X \subset V : |X| = b}{\arg\max} f(X) \; ;
                \end{equation}
            where $f(X) = \frac{|Y|}{|E|}$ with $Y = \{(u, v) \in E : u \in X, v \in V\}$.
            
        \noindent\textbf{Budgeted Max Cut (BMC):}
            Given a graph $G = (V, E)$ and a budget $b$, find a set $X^*$ of $b$ vertices such that the cut set induced by $X^*$ is maximised, i.e.
                \begin{equation}
                    X^* = \underset{X \subset V : |X| = b}{\arg\max} f(X) \; ;
                \end{equation}
            where $f(X) = |\{(u, v) \in E : v \in X, u \in V \setminus X\}|$.
            
        \noindent \textbf{Influence Maximisation (IM):}
            Given a directed weighted graph $G = (V, E)$, a budget $b$, and an information diffusion model $\mathcal{D}$, we are to select a set $X^*$ of $b$ vertices such that the expected spread of influence under $\mathcal{D}$ is maximised, i.e.
                $$X^* = \underset{X \subset V : |X| = b}{\arg\max} \mathbb{E}_\mathcal{D}\left[f(X)\right] \; ;$$
            where $f(X)$ denotes the spread of a set of vertices $X$. We consider the spread of influence under the Independent Cascade diffusion model introduced by \citet{kempe2003maximizing}.
        
        Each problem is tested using eight real-world graphs obtained from the Snap-Stanford repository \cite{snapnets}. For each graph, we randomly sample a percentage of the graphs' edges to form a training graph, with testing then done on the graph formed from the held out edges; the datasets used are summarised in Appendix \ref{sec: dataset information}. All training and test episodes start from different random initial subgraphs. In all experiments, the encoder is trained to identify optimal subgraphs using a budget of $b=100$. Network architectures and hyper-parameters are detailed in Appendix \ref{sec: network architecture and hyper-parameters}.
        
        For our experiments we use $K=4$ classes. The label mapping function $g(x)$ assigns label $1$ if $x \in (0.95, \infty)$ (i.e. if the subgraph is optimal), $2$ if $x \in (0.8, 0.95]$, $3$ if $x \in (0.6, 0.8]$ and $4$ if $x \in [0, 0.6]$; this is done for all graphs with the exception of the Facebook-IM/MVC problem where we only use classes 1-3. We choose a cut-off of $0.95$ for the optimal subgraph in LeNSE to allow for some error when using stochastic solvers. 

        \subsection*{Competing methods}
        
            \begin{table*}[h!]
                \centering
                \begin{tabular}{|c||c|c|c||c|c|c||c|c|c||c|c|c|}
                \hline
                \multirow{2}[4]{*}{\textbf{Graph}} & \multicolumn{3}{c||}{\textbf{LeNSE}} & \multicolumn{3}{c||}{\textbf{GNN-R}} & \multicolumn{3}{c||}{\textbf{GNN-T}} & \multicolumn{3}{c|}{\textbf{GCOMB-P}} \\
                \hline          & Ratio & $P_V$     & $P_E$     & Ratio & $P_V$     & $P_E$     & Ratio & $P_V$     & $P_E$     & Ratio & $P_V$     & $P_E$ \\
                \hline
                        & \multicolumn{12}{c|}{\textbf{Max Vertex Cover}} \\
                \hline
                Facebook & 0.968 (0.009) & 7\%   & 73\%  & 0.998 & 6\%   & 2\%   & 0.607 & 87\%  & 77\%  & 0.927 & 7\%   & 1\% \\
                \hline
                Wiki  & 1.094 (0.000) & 34\%  & 51\%  & 1.003 & 34\%  & 5\%   & 0.371 & 93\%  & 83\%  & 0.990 & 3\%   & 3\% \\
                \hline
                Deezer & 0.979 (0.002) & 75\%  & 94\%  & 0.944 & 75\%  & 66\%  & 0.768 & 99\%  & 99\%  & 0.994 & 13\%  & 3\% \\
                \hline
                Slashdot & 0.979 (0.001) & 69\%  & 83\%  & 0.944 & 66\%  & 30\%  & 0.410 & 99\%  & 95\%  & 1.000 & 2\%   & 10\% \\
                \hline
                Twitter & 0.989 (0.001) & 33\%  & 78\%  & 0.985 & 32\%  & 11\%  & 0.229 & 99\%  & 99\%  & 0.997 & 17\%  & 2\% \\
                \hline
                DBLP  & 0.990 (0.001) & 90\%  & 96\%  & 0.812 & 90\%  & 80\%  & 0.764 & 99\%  & 99\%  & 0.999 & 3\%   & 1\% \\
                \hline
                YouTube & 0.982 (0.002) & 79\%  & 85\%  & 0.417 & 78\%  & 75\%  & 0.317 & 99\%  & 99\%  & 0.998 & 7\%   & 3\% \\
                \hline
                Skitter & 0.976 (0.002) & 70\%  & 84\%  & 0.494 & 70\%  & 69\%  & 0.424 & 99\%  & 99\%  & 0.999 & 10\%  & 2\% \\
                \hline
                      & \multicolumn{12}{c|}{\textbf{Budgeted Max-Cut}} \\
                \hline
                Facebook & 0.960 (0.004) & 7\%   & 67\%  & 0.999 & 8\%   & 3\%   & 0.873 & 89\%  & 70\%  & 0.813 & 95\%  & 89\% \\
                \hline
                Wiki  & 0.981 (0.001) & 39\%  & 63\%  & 0.997 & 30\%  & 12\%  & 0.877 & 97\%  & 96\%  & 0.920 & 96\%  & 91\% \\
                \hline
                Deezer & 0.975 (0.002) & 74\%  & 94\%  & 0.990 & 68\%  & 55\%  & 0.938 & 94\%  & 92\%  & 0.850 & 99\%  & 99\% \\
                \hline
                Slashdot & 0.990 (0.001) & 62\%  & 79\%  & 0.998 & 58\%  & 28\%  & 0.561 & 99\%  & 98\%  & 0.632 & 99\%  & 99\% \\
                \hline
                Twitter & 0.987 (0.001) & 48\%  & 87\%  & 0.991 & 44\%  & 25\%  & 0.820 & 99\%  & 99\%  & 0.628 & 99\%  & 99\% \\
                \hline
                DBLP  & 0.993 (0.000) & 92\%  & 97\%  & 0.875 & 92\%  & 81\%  & 0.656 & 99\%  & 98\%  & 0.646 & 99\%  & 98\% \\
                \hline
                YouTube & 0.987 (0.001) & 79\%  & 84\%  & 0.613 & 74\%  & 74\%  & 0.753 & 99\%  & 99\%  & 0.536 & 99\%  & 97\% \\
                \hline
                Skitter & 0.974 (0.004) & 71\%  & 83\%  & 0.502 & 71\%  & 60\%  & 0.407 & 99\%  & 99\%  & 0.427 & 99\%  & 99\% \\
                \hline
                      & \multicolumn{12}{c|}{\textbf{Influence Maximisation}} \\
                \hline
                Facebook & 0.979 (0.002) & 9\%   & 70\%  & 1.006  & 9\%   & 6\%   & 0.886 & 91\%  & 88\%  & 0.951 & 73\%  & 63\% \\
                \hline
                Wiki  & 0.960 (0.002) & 51\%  & 71\%  & 0.964 & 49\%  & 78\%  & 0.973 & 96\%  & 95\%  & 0.969 & 90\%  & 81\% \\
                \hline
                Deezer & 0.972 (0.003) & 76\%  & 94\%  & 0.935 & 76\%  & 86\%  & 0.775 & 98\%  & 99\%  & 0.805 & 95\%  & 97\% \\
                \hline
                Slashdot & 0.966 (0.003) & 77\%  & 90\%  & 0.966 & 74\%  & 71\%  & 0.951 & 98\%  & 95\%  & 0.966 & 98\%  & 93\% \\
                \hline
                Twitter & 0.966 (0.001) & 40\%  & 88\%  & 0.988 & 26\%  & 12\%  & 0.921 & 98\%  & 97\%  & 0.920 & 98\%  & 97\% \\
                \hline
                DBLP  & 0.969 (0.002) & 89\%  & 96\%  & 0.935 & 89\%  & 88\%  & 0.844 & 99\%  & 98\%  & 0.863 & 99\%  & 99\% \\
                \hline
                YouTube & 0.971 (0.001) & 75\%  & 81\%  & 0.918 & 75\%  & 75\%  & 0.806 & 94\%  & 99\%  & 0.933 & 99\%  & 99\% \\
                \hline
                Skitter & 0.983 (0.002) & 78\%  & 85\%  & 0.919 & 69\%  & 58\%  & 0.883 & 99\%  & 99\%  & 0.883 & 99\%  & 99\% \\
                \hline
                \end{tabular}
                \caption{Results for LeNSE, GNN-R, GNN-T and GCOMB-P. The ratio reported is the ratio from Equation \eqref{eq: objective} and $P_V, P_E$ denote the percentage of vertices and edges, respectively, pruned from the graph. The results for LeNSE are averaged over 10 random initial subgraphs with the standard errors for the ratio given in the brackets.}
                \label{tab: main results}
            \end{table*}
        
            We compare LeNSE to a GNN vertex classifier and the pruning method described in GCOMB \cite{manchanda2020gcomb} which we denote as GCOMB-P (see Appendix \ref{sec: gcomb pruning} for a more detailed overview). The GNN classifier uses the same architecture as LeNSE's encoder except with no coarsening layers. The classifier is trained to predict the probability that a vertex is one of the $b$ solution vertices. At each epoch, we randomly sample $b$ vertices that are \textit{not} solution vertices, and evaluate the loss on these sampled vertices and the $b$ sampled vertices; this avoids the loss being dominated by the non-solution vertices. Using the trained classifier, we used the predicted probabilities to rank the vertices. We choose two different thresholds using the predicted probabilities. GNN-R: assuming that the subgraph returned by LeNSE has $k$ vertices, we prune from the graph the bottom $|V| - k$ ranked vertices; GNN-T: we remove all vertices with predicted probability less than 0.5. 
                
        \subsection*{Experimental results \label{sec: experimental results}}
            
            We look to assess LeNSE's ability to prune a graph, and the quality of solution that can be found on the pruned graph. To that end, we report 2 main metrics: 1) the number of vertices/edges in the final subgraph; 2) the final ratio achieved, i.e. Equation \eqref{eq: objective}. Note that when reporting Equation \eqref{eq: objective}, the set $X^*$ found by the heuristic on the test graph is used. In Table \ref{tab: main results} we report these metrics for LeNSE and the competing methods. 
            
            We can see that for each problem and dataset LeNSE is able to find an optimal subgraph (i.e. ratio greater than 0.95). Interestingly, we see that for Wiki-MVC the heuristic was able to find a \textit{better} solution on the subgraph than it did on the original graph. Note that this is possible because the greedy heuristic used for MVC has a 60\% approximation ratio, and so with less noise the heuristic is able to find a better solution than on the entire graph.
            
            In terms of the amount of pruning done, we see that LeNSE is consistently able to prune large amounts of edges from the graphs. In most instances we also see that LeNSE prunes large portions of vertices from the graph; in particular we note that for the 3 larger graphs we are able to prune 70+\% in all 3 of the problems. However, as the complexity of many heuristics rely on the number of edges as well as vertices -- as is the case with IMM and the MIP formulation of BMC -- pruning edges is equally important as pruning vertices for reducing the run time of heuristics.
                
            For the competing methods, we can see that whilst GCOMB-P performs well at achieving a close-to optimal ratio for IM and MVC, it performs worse in the BMC problem. This suggests that a hand-crafted pruning method may not be as effective at generalising to multiple problems compared to one which is learnt, such as in LeNSE. Further, GCOMB-P prunes very little of the graphs in the MVC problem which is likely why the ratios are consistently close to optimal. The converse can be said for GCOMB-P in the BMC problems, where it prunes too much of the graphs and hence gives lower ratios. 
            
            The GNN-R method obtains similar ratios to LeNSE for the smaller datasets, but on the larger datasets (e.g. DBLP, YouTube, Skitter) the performance starts to degenerate, suggesting that the GNN classifier is unable to generalise when the test graph is much larger than the training graph. Further, we note that with the exception of Wiki-IM this approach never prunes as many edges as LeNSE. The GNN-T approach achieves a poor ratio across all MVC and BMC datasets due to the over-pruning of the graphs. The performance is improved somewhat in the IM problem, but it still does not achieve ratios close to that obtained by LeNSE.
            
            Further to these results, in Figure \ref{fig: traceplot example} we provide a visual demonstration of the policy learnt by LeNSE for the Wiki-BMC graph. In plot (A) we provide an example trajectory taken by the agent. The agent starts in a region close to the worst class of subgraphs and sequentially makes progress by moving closer to the optimal subgraphs after each action. In plot (B) we show the (scaled) distance per episode time step of the subgraph from the goal. The distance is averaged over 10 randomly initialised episodes, and we can see that the mean distance tends towards 0 as the episode goes on, with the confidence interval becoming tighter.
            
            \begin{figure*}[tp]
                  \centering
                  \includegraphics[width=\linewidth]{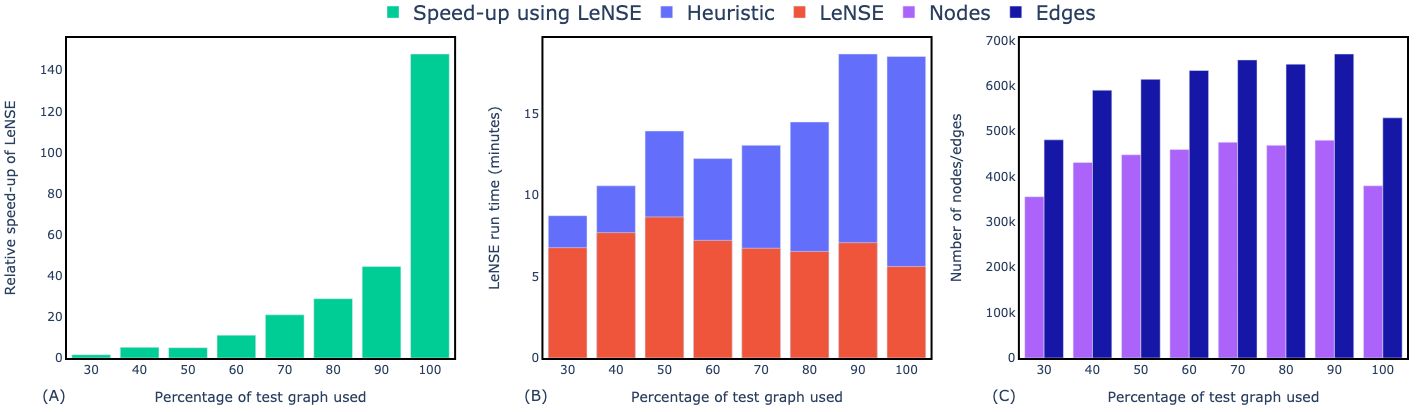}
                  \caption{(A): The relative speed-up of using LeNSE vs. using the heuristic on the entire graph. (B): The time taken, in minutes, for LeNSE to run. We have broken down the run time to demonstrate the time taken for LeNSE to find a subgraph (red bars) and the time taken to run the heuristic on the subgraphs (blue bars). (C): The number of vertices/edges in the final subgraphs returned by LeNSE. In each plot we vary the percentage of the test graph used to demonstrate how the performance of LeNSE changes as the size of the graph increases. These experiments were all performed on the Talk-IM graph.}
                  \label{fig: scalability plot}
            \end{figure*}
                
        \subsection*{Scalability study}
        
            In Figure \ref{fig: scalability plot} we demonstrate the performance of LeNSE on the large Talk graph which has 2.3m vertices and 4.8m edges. In particular, we show how the performance varies as we increase the percentage of the test graph's edges used. We report the relative speed-up of using LeNSE vs. using the heuristic on the entire graph, the run time of LeNSE and the number of vertices/edges in the subgraphs returned by LeNSE -- these are all averaged over 10 random episodes. The experiments were carried out for the IM problem and we note that all the heuristic achieved a ratio greater than 0.95 on all of the subgraphs. 
            
            We can see from the Figure that despite the test graph size increasing, the time required by LeNSE to find a subgraph remains almost constant. This is due to the fact that the amount of pruning done by LeNSE increases as the size of the graph increases, and so the subgraph size passed through the encoder does not increase linearly with the size of the test graph. Importantly, we show that using LeNSE leads to more than a 140 time speed-up as opposed to using the heuristic on the full graph.

        \subsection*{Multi-budget scores} 
        
            We assess whether heuristics can also obtain (close-to) optimal results for budgets lower than that which the encoder is trained for in the pre-processing phase. As mentioned, in our experiments the encoder is trained to identify an optimal subgraph for a budget of 100. We now test whether the subgraphs returned by LeNSE can also be used to obtain an optimal score for various budgets lower than 100. Specifically, we report the ratio achieved when looking to find a solution for the following budgets: 1, 10, 25, 50 and 75. 
            
            The results for the three problems can be found in Table \ref{tab: multi budget results} in Appendix \ref{sec: multi budget appendix}. In the majority of cases the subgraphs provided by LeNSE are also optimal subgraphs for budgets lower than that which LeNSE was trained for. The exceptions are the Facebook-IM, DBLP-IM and Skitter-IM/MVC graphs for a budget of one. Despite these four instances, the evidence would suggests that a viable training option for LeNSE is to train for the largest budget that will ever be required, and it should follow that a heuristic can be used to find an optimal solution for all smaller budgets.
        
\section{Conclusion}
    \label{sec: Conclusion}
    
    We have introduced LeNSE, a novel graph pruning algorithm that finds an optimal subgraph containing the solution of a CO problem. Rather than being a single algorithm, LeNSE is a general framework that can be used in combination with any existing heuristic of choice for large-scale problems. In all the settings that we have investigated using real-world graph datasets, we found that training could be done on some small percentage of the graph, being able to scale up to the held out portion of the graph without any performance degradation. We also found that the subgraphs found by LeNSE were substantially smaller than the original graph, with more than 90\% of the graphs vertices and edges being removed in the best case. 
    
    Further to this, we also demonstrated that the subgraphs found by LeNSE were, in most cases, optimal subgraphs for budgets lower than that which the encoder was trained to recognise. We compared LeNSE to two baseline methods where we found that LeNSE was able to either provide a (much) better ratio than the comparisons, or prune a significant amount more of the graph whilst still finding an optimal subgraph. As well as this, LeNSE was able to consistently perform well across all problems. Finally, we demonstrated the benefits of using LeNSE on a graph with nearly $5$ million edges, showing that LeNSE achieved speed up of more than 140 times compared to not doing any pruning.
    
    Potential avenues for future work include alternative graph modification operations, as in this work we only consider the replacement of a vertex with one of its neighbours and include all of the one-hop neighbours of some vertex subset in the subgraph. We also plan to further investigate LeNSE's performance on other CO problems besides the budget-constrained versions presented here.
    
\newpage
\bibliography{ref}
\bibliographystyle{icml2022}

\newpage
\onecolumn
\appendix

    \section{Dataset generation}
        
        \label{sec: dataset generation}
    
        Algorithm \ref{alg: dataset generation} describes the procedure used to generate the dataset consisting of labelled subgraphs which are used to learn a discriminative subgraph embedding. Here, we recall that $G_T$ represents the train graph, and so the generated dataset will consist of subgraphs of the smaller train graph. To generate the training dataset, we sample random subgraphs and compute the corresponding label, which are assigned using a function $g(\cdot)$. We ensure a balanced dataset (i.e. an equal number of subgraphs for each one of the $K$ classes); this is achieved by ensuring that some fraction of the $b$ solution vertices, which we assume to be known on the train-graph, are present in the subgraph so that we achieve the desired score.
        
            \begin{algorithm}
                \caption{Dataset generation \label{alg: dataset generation}}
                \begin{algorithmic}[1]
                    \REQUIRE Train graph $G_T = (V_T, E_T)$, Heuristic Solver $\mathcal{H}$, Heuristic score $f(\mathcal{H}(G))$, Dataset size $N$, label mapping function $g(\cdot)$, Fixed vertex size $M$
                    \STATE $D \leftarrow \emptyset $
                    \FOR{$i \mbox{ in } 1, 2, ..., N$}
                        \STATE $X \sim \mathcal{U}(V)$ such that $|X| = M$
                        \STATE $S = \xi(X)$
                        \STATE Compute ratio $r = f(\mathcal{H}(S)) / f(\mathcal{H}(G))$
                        \STATE Assign label $y = g(r)$
                        \STATE $D \leftarrow D \cup (S, y)$
                    \ENDFOR
                \end{algorithmic}
            \end{algorithm}
            
    \section{Dataset information}
        \label{sec: dataset information}
        In Table \ref{tab: dataset information} we list the graphs used in our experiments and the size of the respective training and testing splits. In addition to the number of edges used for training, we also indicate what percentage of the original graphs edges were used. Note that encoder training and DQN training are all carried out on the train graph, and then tested on the held out test graph.
        
        \begin{table}
          \centering
            \begin{tabular}{|c|r|r|r|r|}
                \hline
                \multirow{2}[4]{*}{Graph Name} & 
                    \multicolumn{2}{|c|}{Train Size} & 
                    \multicolumn{2}{|c|}{Test Size} \\ 
                    \cline{2-5} & \multicolumn{1}{|c|}{Vertices} & \multicolumn{1}{|c|}{Edges} & \multicolumn{1}{|c|}{Vertices} & \multicolumn{1}{|c|}{Edges} \\
                \hline
                Facebook & 3,822 & 26,470 (30\%) & 3,998 & 61,764 \\
                Wiki & 4,895 & 31,106 (30\%) & 6,349 & 72,583 \\
                Deezer & 48,775 & 149,460 (30\%) & 53,555 & 348,742\\
                Slashdot (U) & 47,558 & 140,566 (30\%) & 67,677 & 327,988 \\
                Slashdot (D) & 48,976 & 154,972 (30\%) & 68,292 & 361,603 \\
                Twitter (U) & 55,911 & 134,231 (10\%) & 80,779 & 1,208,079 \\
                Twitter (D) & 58,981 & 176,814 (10\%) & 80,776 & 1,591,335 \\
                DBLP & 36,889 & 42,467 (04\%) & 314,818 & 1,007,399 \\
                YouTube & 125,607 & 185,566 (06\%) & 1,094,439 & 2,802,058 \\
                Skitter & 77,179 & 115,170 (01\%) & 1,694,024 & 10,980,128 \\
                Talk & 120,443 & 181,843 (04\%) & 2,329,147 & 4,839,567 \\
            \hline
            \end{tabular}
            \caption{The real-world graphs used to perform our experiments. For each graph, we provide the corresponding size of the training and testing sets;  in brackets we indicate the percentage of edges taken from the original graph for training purposes with the remaining edges being used to form the test graph. For Slashdot and Twitter, the size of the original graph varied when we treated it as undirected/directed, so (U) refers to undirected graphs and (D) denotes the directed one. All datasets can be found at \url{http://snap.stanford.edu/data/index.html}.}
            \label{tab: dataset information}
        \end{table}
            
    \section{Network architectures and hyper-parameters}
        \label{sec: network architecture and hyper-parameters}
        
        The encoder must be able to provide meaningful graph level embeddings, as opposed to vertex/edge embeddings. Therefore we combine convolutional layers with \textit{graph coarsening} layers. The convolutional layer learns vertex features which are then used by the coarsening layer to reduce the size of the graph. Repeatedly applying such layers leads to a single vectorial representation of the input graph. In particular, our encoder architecture consists of GraphSAGE layers \cite{hamilton2017inductive} to learn the vertex features and $k$-pooling layers \cite{cangea2018towards} to coarsen the graph. 
        
        Our encoder architecture has 3 layers, each one consisting of a GraphSAGE layer with ReLU activation followed immediately by a $k$-pooling layer. The output is a single vector representation of the input subgraph which is then passed through a fully connected linear layer to map to the desired dimension -- the full architecture is summarised in Figure 1 of \citet{cangea2018towards}. For the MVC and BMC problem we use as raw vertex features the number of neighbours and the eigenvector centrality \cite{newman2008mathematics}, and for the IM problem we additionally use the sum of the outgoing edge-weights of a vertex -- note these are all computed relative to the original graph, not the subgraphs. Note that we normalise the number of vertices and the outgoing edge weights by min-max.
        
        Unlike the more commonly used DQN architecture which takes in the state and outputs a Q-value for each possible action, our DQN will take in the state and action being considered and output a single Q-value. This is necessary as our action space changes at each step of an episode based on which action tuples are in $\mathcal{A}_t$. The DQN architecture consists of 3 independent fully connected layers where the output of each is concatenated into a single vector before being passed through 3 further linear layers. The number of hidden units in all fully connected layers is 128 and all have ReLU activations, except for the final layer which is linear as we want to map to $\mathbb{R}$ for the Q-Value. Formally, our DQN is a function $Q: \mathbb{R}^{d} \times \mathbb{R}^{d'} \times \mathbb{R}^{d'} \rightarrow \mathbb{R}$ where $d$ is the subgraph embedding dimensionality and $d'$ is the dimensionality of the vertex embeddings from the first GraphSAGE layer in the encoder. We use this first GraphSAGE layer (i.e. before any pooling) to obtain vertex embeddings when passing the actions (vertices) through the DQN. 
        
        The weights of both the encoder and the DQN are optimised with the Adam optimiser \cite{kingma2014adam} and both have a learning rate of 0.001. The encoder batch size and experience replay batch size are both 128. We use a pooling ratio of $k = 0.8$ and temperature hyper-parameter of $\tau = 0.1$ in the encoder. Finally, when training the DQN we initialise the exploration parameter $\epsilon$ to be equal to one and decay it after each random action at a rate of 0.9995 down to a minimum value of 0.01. We also have a max replay memory size of 25,000, a discount factor of 0.995 and a target network which is updated after each experience replay update using polyak averaging with a parameter of 0.0025. Note that during training we use a different episode length than at test time, both of which we report, along with the remaining hyper-parameters that vary per dataset, in Table \ref{tab: parameters used}. Note that we only report parameters for the Talk-IM graph as we do not perform any experiments using the Talk-BMC/MVC graphs.

            \begin{table*}[h!]
              \centering
                \begin{tabular}{|c|c|c|c|c|c|c|c|c|c|c|}
                \hline
                \multirow{2}[4]{*}{\textbf{Graph}} & \multicolumn{1}{c|}{$M$} & \multicolumn{1}{c|}{$n$} & \multicolumn{1}{c|}{$d'$} & \multicolumn{1}{c|}{$d$} & \multicolumn{1}{c|}{$T_{train}$} &
                \multicolumn{1}{c|}{$N'$} & $\alpha$ & \multicolumn{1}{c|}{$c$} & $\beta$ & $T_{test}$ \\
                \hline
                & \multicolumn{10}{c|}{\textbf{MVC}} \\
                \hline
                Facebook &              300  & 100 & 30    & 10    &            2,000  & 10   & 0     & 20    & 50 & 2,000\\
                \hline
                Wiki  &              300  & 100 & 75    & 25    &            2,000  & 10     & 0.1     & 20    & 3 & 300 \\
                \hline
                Deezer &              500  & 200 & 50    & 10    &            2,000  & 10    & 0     & 20    & 15 & 300 \\
                \hline
                Slashdot &              500  & 200 & 30    & 15    &            2,000  & 10    & 0.1     & 20    & 20 & 300\\
                \hline
                Twitter &           1,000  & 100 & 40    & 10    &            1,500  & 10     & 0.1   & 2     & 20 & 750 \\
                \hline
                DBLP  &           1,000  & 100 & 40    & 10    &            1,500  & 10   & 0.1   & 2     & 20 & 1,000 \\
                \hline
                Youtube &           1,250  & 100 & 75    & 25    &            1,500  & 10     & 0.1   & 2     & 30 & 500 \\
                \hline
                Skitter &              750  & 250 & 75    & 20    &            1,000  & 10    & 0.05  & 2     & 30 & 500 \\
                \hline
                      & \multicolumn{10}{c|}{\textbf{Budgeted Max-Cut}} \\
                \hline
                Facebook &              300 & 250  & 40    & 10    &            2,000  & 10     & 0.05     & 20    & 20 & 2,000 \\
                \hline
                Wiki  &              300  & 250 & 30    & 10    &            2,000  & 10    & 0.05     & 20    & 25 & 100\\
                \hline
                Deezer &              500 & 250  & 30    & 10    &            2,000  & 10    & 0.05     & 20    & 50 & 300 \\
                \hline
                Slashdot &              500 & 250  & 30    & 15    &            2,000  & 10    & 0.05     & 20    & 50 & 300\\
                \hline
                Twitter &           1,000  & 100 & 40    & 10    &            2,000  & 10     & 0.5   & 2     & 20 & 250 \\
                \hline
                DBLP  &           1,000  & 100 & 30    & 10    &            2,000  & 10    & 0.5   & 2     & 50 & 1,000 \\
                \hline
                Youtube &           1,000  & 200 & 75    & 25    &            1,500  & 10    & 0.5   & 2     & 5 & 400 \\
                \hline
                Skitter &              750  & 500 & 75    & 25    &            1,000  & 10    & 0.05  & 2     & 30 & 500\\
                \hline
                      & \multicolumn{10}{c|}{\textbf{Influence Maximisation}} \\
                \hline
                Facebook &              300 & 100  & 30    & 10    &            5,000  & 10    & 0     & 20    & 50 & 5,000 \\
                \hline
                Wiki  &              300  & 100 & 50    & 15    &            2,000  & 10    & 0     & 20    & 50 & 200 \\
                \hline
                Deezer &              500  & 100 & 30    & 10    &            2,000  & 10    & 0     & 20    & 15 & 1,500\\
                \hline
                Slashdot &              500  & 100 & 50    & 10    &            2,000  & 10    & 0     & 20    & 15 & 300 \\
                \hline
                Twitter &           1,000  & 100 & 40    & 10    &            1,500  & 10     & 0.5   & 2     & 20 & 400 \\
                \hline
                DBLP  &           1,000  & 100 & 40    & 10    &            1,500  & 10     & 0.5   & 2     & 20 & 1,000 \\
                \hline
                Youtube &              750  & 100 & 75    & 25    &            1,500  & 10    & 0.1   & 2     & 20 & 500 \\
                \hline
                Skitter &              750  & 250 & 75    & 20    &               500  & 25    & 0.05  & 2     & 10 & 500\\
                \hline
                Talk &                 750 & 400 & 75 & 25 & 500 & 25 & 0.1 & 2 & 10 & 100 \\
                \hline
                \end{tabular}%
              \caption{The hyperparameters used when training LeNSE. $M$ denotes the size of the vertex subset we use to induce the subgraphs throughout training, $n$ denotes the number of samples per class in the subgraph dataset, $N'$ is the number of training episodes, $d'$ corresponds to the dimensionality of the vertex embeddings for the GraphSAGE layers in the encoder whilst $d$ is the final embedding dimension of the encoder, $c$ denotes the frequency with which we perform SGD on the parameters of the DQN as per Algorithm \ref{alg: LeNSE}, $\beta$ is the scaling factor in the rewards and $T_{train}, T_{test}$ denote the length of the train and test episodes, respectively.}
              \label{tab: parameters used}%
            \end{table*}%

    \section{Multi budget test results}
        
        \label{sec: multi budget appendix}
    
        Table \ref{tab: multi budget results} contains the results for the multi budget test results referenced in the main paper. We can see that all the ratios, except for Facebook-IM, DBLP-IM and Skitter-MVC/IM, are ratios we would expect from an optimal subgraph. In particular, the ratios for the deterministic solvers in MVC and BMC are all close to 1.
    
            \begin{table*}[h!]
              \centering
                \begin{tabular}{|c|c|c|c|c|c|c|}
                \hline
                \multirow{2}[4]{*}{\textbf{Graph}} & 
                    \multicolumn{6}{|c|}{\textbf{Budget}} \\
                    \cline{2-7} & 1 & 10 & 25 & 50 & 75 & 100 \\
                \hline
                & \multicolumn{6}{c|}{\textbf{Max Vertex Cover}} \\
                \hline
                Facebook & 1.00 & 1.00 & 0.99 & 0.97 & 0.97 & 0.97\\
                Wiki  & 1.00 & 1.00 & 1.00 & 1.00 & 1.04 & 1.09 \\
                Deezer & 1.00 & 1.00 & 1.00 & 0.99 & 0.99 & 0.98 \\
                Slashdot & 1.00 & 1.00 & 1.00 & 1.00 & 0.99 & 0.98 \\
                Twitter & 1.00 & 1.00 & 1.00 & 1.00 & 1.00 & 0.99 \\
                DBLP  & 1.00 & 1.00 & 1.00 & 1.00 & 0.99 & 0.99 \\
                YouTube & 1.00 & 1.00 & 1.00 & 1.00 & 0.99 & 0.98 \\
                Skitter & 0.85 & 0.98 & 0.99 & 0.99 & 0.99 & 0.98 \\
                \hline
                & \multicolumn{6}{c|}{\textbf{Budgeted Max-Cut}} \\
                \hline
                Facebook & 1.00 & 1.00 & 0.99 & 0.98 & 0.98 & 0.96\\
                Wiki  & 1.00 & 1.00 & 1.00 & 1.00 & 1.00 & 0.98 \\
                Deezer & 1.00 & 1.00 & 1.00 & 0.99 & 0.99 & 0.98 \\
                Slashdot & 1.00 & 1.00 & 1.00 & 1.00 & 1.00 & 0.99 \\
                Twitter & 0.98 & 1.00 & 1.00 & 1.00 & 0.99 & 0.99 \\
                DBLP  & 1.00 & 1.00 & 1.00 & 1.00 & 1.00 & 0.99 \\
                YouTube & 1.00 & 1.00 & 1.00 & 1.00 & 0.99 & 0.99 \\
                Skitter & 0.96 & 1.00 & 1.00 & 0.99 & 0.99 & 0.97 \\
                \hline
                & \multicolumn{6}{c|}{\textbf{Influence Maximisation}} \\
                \hline
                Facebook & 0.86 & 0.99 & 0.98 & 0.99 & 0.99 & 0.98 \\
                Wiki  & 1.00 & 0.98 & 0.99 & 1.00 & 0.98 & 0.96 \\
                Deezer & 1.00 & 0.97 & 0.97 & 0.97 & 0.97 & 0.97 \\
                Slashdot & 1.00 & 0.99 & 0.99 & 0.98 & 0.98 & 0.98 \\
                Twitter & 1.00 & 0.97 & 0.96 & 0.96 & 0.96 & 0.96 \\
                DBLP  & 0.86 & 0.96 & 0.95 & 0.97 & 0.96 & 0.97 \\
                YouTube & 1.00 & 0.96 & 0.98 & 0.98 & 0.97 & 0.97 \\
                Skitter & 0.77 & 0.97 & 0.98 & 0.98 & 0.98 & 0.98 \\
                \hline
                \end{tabular}
                \caption{Results showing the ratio achieved for the three CO problems, averaged over 10 runs, for budgets lower than that which the encoder was trained to attain.}
                \label{tab: multi budget results}
            \end{table*}
            
    \section{GCOMB-based graph pruning}
        \label{sec: gcomb pruning}
        
        The methodology presented in \citet{manchanda2020gcomb} can be broken down into two parts -- a pruning phase and a Q-learning phase. The pruning is based on a vertex ranking approach. For a train graph $G = (V, E)$ and budget of $b$, the vertices are initially sorted into descending order based on the outgoing edge-weight (or degree in an unweighted graph) and $rank(v)$ denotes the position of vertex $v$ in this ordered list. A stochastic solver is used $L$ times to obtain $L$ (different) solution sets $\{A^{(i)}\}_{i=1}^L$ of size $b$. If we define $A^{(i)}_{:k}$ to be the first $k$ vertices added to solution set $A^{(i)}$, then for all budgets $b' \leq b$ we define $r_{b'} = \max_{v \in \cup_{i} A^{(i)}_{:k}} rank(v)$ to be the highest rank of all vertices amongst the first $b'$ vertices in each of the $L$ solution sets. A linear interpolator is then used to fit these $(b', r_{b'})$ pairs. Now, for a budget $\tilde{b}$, the test graph is pruned by using the linear interpolator to predict a rank $\hat{r}_{\tilde{b}}$ and remove all vertices in the test graph with rank greater than $\hat{r}_{\tilde{b}}$, where the rankings of the test graph and calculated in the same way as for the train graph. Note that when training the linear interpolator the rankings and budgets are normalised by the proportion of vertices in the graph to enable generalisation to different graphs. In our experiments, the stochastic solvers that we use for the 3 problems are IMM \cite{tang2015influence} for IM, the heuristic introduced in \citet{hassin2000approximation} for BMC and for MVC we use the probabilistic-greedy that was introduced in the GCOMB paper. For the Q-learning phase of GCOMB, a simple Q-learning algorithm is deployed on the pruned graph. This is in contrast to LeNSE, where we use an existing heuristic. 
        
    \section{Loss Comparisons}

        \label{sec: loss ablation study}
    
        \begin{table}[H]
                \centering
                    \begin{tabular}{|c|c|c|c|}
                        \hline
                        \multirow{2}[4]{*}{\textbf{Graph}} & \multicolumn{3}{c|}{\textbf{Loss}} \\
                        \cline{2-4}  & InfoNCE & Cross-Entropy & Ordinal \\
                        \hline
                        & \multicolumn{3}{c|}{\textbf{Max Vertex Cover}} \\
                        \hline
                        Facebook & 0.968 & 0.907 & 0.880 \\
                        \hline
                        Wiki  & 1.094 & 0.899 & 0.931 \\
                        \hline
                        & \multicolumn{3}{c|}{\textbf{Budgeted Max-Cut}} \\
                        \hline
                        Facebook & 0.960 & 0.647 & 0.650 \\
                        \hline
                        Wiki  & 0.981 & 0.391 & 0.603 \\
                        \hline
                        & \multicolumn{3}{c|}{\textbf{Influence Maximisation}} \\
                        \hline
                        Facebook & 0.979 & 0.663 & 0.871 \\
                        \hline
                        Wiki  & 0.960 & 0.895 & 0.914 \\
                        \hline
                    \end{tabular}%
                \caption{Comparison of the average ratios from Equation \eqref{eq: objective} achieved when using the different loss functions.}
                \label{tab: loss ablation}%
            \end{table}%
            
        We compare the InfoNCE loss to the standard cross entropy loss as well as to an ordinal classification based loss. We choose to compare to an ordinal based loss function as there is an ordinal structure to the classes. For example, it would be more acceptable to misclassify a class 4 subgraph as a class 3 subgraph than it would be to classify it as a class 1 subgraph. We choose the ordinal classification approach introduced in \citet{cheng2008neural}. With $K$ classes, a subgraph with class $i \leq K$ is assigned an ordinal target vector $y$ where $y_j=1$ for $j \leq i$, and 0 otherwise. The output layer uses a sigmoid activation so that the predicted vector $\hat{y}$ has individual values in the range $(0, 1)$ and the network is trained to minimise the mean squared error between the $\hat{y}$ and $y$. To make the comparisons fair we use the same encoder (i.e. same architecture) for each loss function, where the ordinal and cross entropy have the additional layer to map from the embedding space to the prediction vector as required. 
        
\begin{figure}[H]
                \centering
                \includegraphics[width=\linewidth]{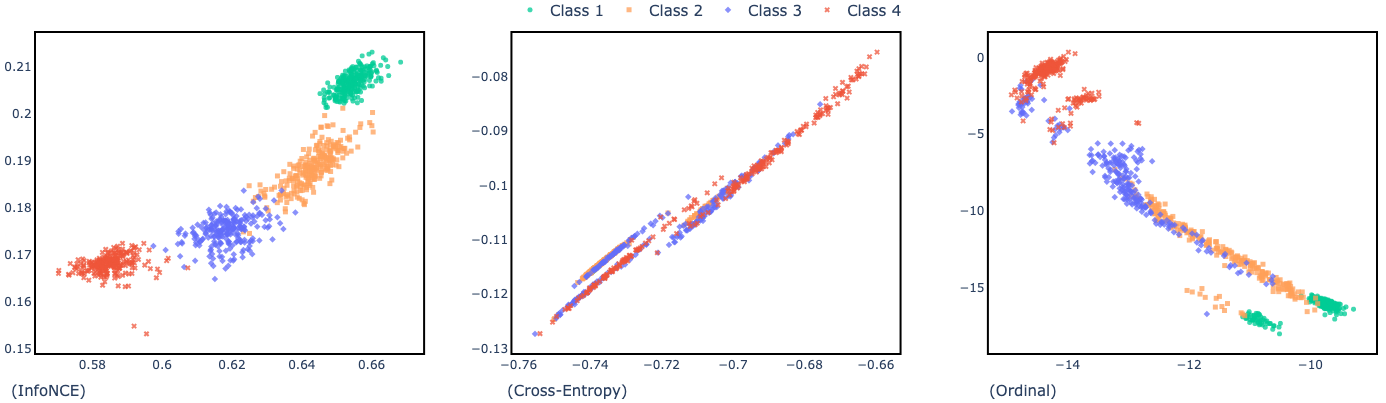}
                \caption{Example of the embeddings provided for the Wiki-BMC graph by the InfoNCE loss, cross entropy loss and ordinal loss. Note that autoencoders were used for dimensionality reduction to map from the original embedding space to the plotted 2-dimensional space.}
                \label{fig: joint embeddings}
            \end{figure}

\begin{figure}[H]
                \centering
                \includegraphics[width=\linewidth]{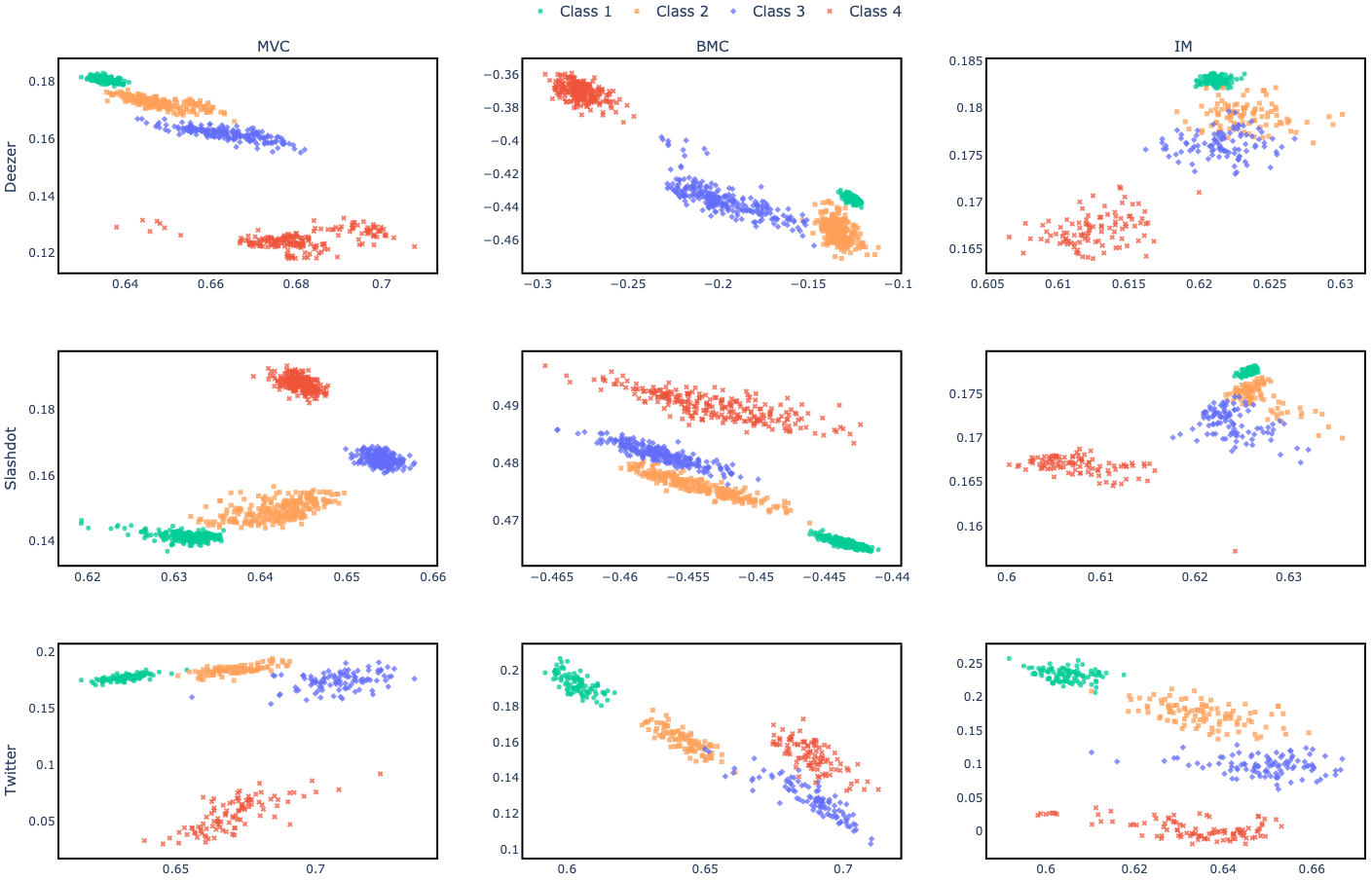}
                \caption{Example of embeddings provided by the InfoNCE loss for various graph/problem combinations. Note that autoencoders were used for dimensionality reduction to map from the original embedding space to the plotted 2-dimensional space.}
                \label{fig: big embeddings}
            \end{figure}

        In Figure \ref{fig: joint embeddings} we present example embeddings using the 3 different loss functions for the Wiki-BMC graph, and in Table \ref{tab: loss ablation} we present the corresponding average ratios obtained by LeNSE when using embeddings trained with the respective losses. We can see from the table that the variation is much higher when using cross entropy and ordinal trained encoders. This is likely due to what we see in Figure \ref{fig: joint embeddings} where the embeddings provided are not as well separated as when using the InfoNCE loss. In particular, the cross entropy embeddings have no clear structure, with class 3 and 4 subgraphs being embedded in the same location as class 1 and 2 (hence why they cannot be seen on the Figure) -- this is likely the cause for LeNSE achieving a ratio of just 0.391. Interestingly, the ordinal embeddings in the Figure highlight why the InfoNCE loss is useful as we can see that for class 1 there are two separate clusters, including some overlap with class 2. This highlights why uni-modality is useful as LeNSE obtained a ratio of only 0.603 using this embedding. This can be explained by the fact that the goal would have been at the centre of the two clusters which does not necessarily correspond to a good region of the embedding. In contrast to this, we see in Figure \ref{fig: big embeddings} that all the embeddings provided by the InfoNCE loss have uni-modal clusterings with the monotonic ordering of the classes maintained, with an exception in the Twitter-BMC graph where classes 3 and 4 are well separated but are approximately a similar distance from the class 1 embeddings.
        
\end{document}